\documentclass[conference]{IEEEtran}
\IEEEoverridecommandlockouts
\usepackage{cite}
\usepackage{amsmath,amssymb,amsfonts}
\usepackage{algorithmic}
\usepackage{graphicx}
\usepackage{textcomp}
\usepackage{xcolor}
\usepackage{algorithm}
\usepackage{algorithmic}
\usepackage{multirow}
\usepackage{subfigure}
\usepackage{setspace}
\usepackage{amsthm}
\usepackage{url}
\usepackage{booktabs}

\newcommand{\model}{SECoExpan}

\def\BibTeX{{\rm B\kern-.05em{\sc i\kern-.025em b}\kern-.08em
    T\kern-.1667em\lower.7ex\hbox{E}\kern-.125emX}}

\newcommand{\abs}[1]{\mathopen| #1 \mathclose|}	
\DeclareMathOperator*{\argmax}{arg\,max}
\newcommand{\mquote}[1]{{``\textit{#1}''}}

\definecolor{myred}{rgb}{0.85, 0.0, 0.0}
\newcommand*\red{\color{myred}}

\begin{document}

\title{Entity Set Co-Expansion in StackOverflow}

\author{
\IEEEauthorblockN{Yu Zhang$^{1*}$\thanks{$^*$Equal Contribution.}, Yunyi Zhang$^{1*}$, Yucheng Jiang$^1$, Martin Michalski$^1$, \\ Yu Deng$^2$, Lucian Popa$^3$, ChengXiang Zhai$^1$, Jiawei Han$^1$}
\IEEEauthorblockA{$^1$Department of Computer Science, University of Illinois at Urbana-Champaign, Urbana, IL, USA \\
$^2$IBM Thomas J. Watson Research Center, Yorktown Heights, NY, USA \\
$^3$IBM Almaden Research Center, San Jose, CA, USA \\
\{yuz9, yzhan238, yj17, martinm6, czhai, hanj\}@illinois.edu, \ \ \  \{dengy, lpopa\}@us.ibm.com}
}

\maketitle

\begin{spacing}{0.96}
\begin{abstract}
Given a few seed entities of a certain type (e.g., \textsc{Software} or \textsc{Programming Language}), entity set expansion aims to discover an extensive set of entities that share the same type as the seeds. Entity set expansion in software-related domains such as StackOverflow can benefit many downstream tasks (e.g., software knowledge graph construction) and facilitate better IT operations and service management. Meanwhile, existing approaches are less concerned with two problems: (1) How to deal with multiple types of seed entities simultaneously? (2) How to leverage the power of pre-trained language models (PLMs)? Being aware of these two problems, in this paper, we study the entity set co-expansion task in StackOverflow, which extracts \textsc{Library}, \textsc{OS}, \textsc{Application}, and \textsc{Language} entities from StackOverflow question-answer threads. During the co-expansion process, we use PLMs to derive embeddings of candidate entities for calculating similarities between entities. Experimental results show that our proposed \textsc{\model} framework outperforms previous approaches significantly.

\end{abstract}

\begin{IEEEkeywords}
set expansion, entity extraction, StackOverflow
\end{IEEEkeywords}

\section{Introduction}
The task of entity set expansion \cite{rong2016egoset,shen2017setexpan} aims to enrich a small set of seed entities (e.g., ``\textit{Java}'', ``\textit{C++}'', and ``\textit{PHP}'') by extracting other entities belonging to the same type (e.g., ``\textit{Python}'', ``\textit{SQL}'', and ``\textit{JavaScript}'' that are also programming languages) from a large corpus. Previous studies have shown the benefit of entity set expansion to a wide range of downstream applications, such as named entity recognition \cite{wang2019distantly}, taxonomy construction \cite{shen2018hiexpan}, and text classification \cite{zhang2019higitclass}.

While existing approaches demonstrate their effectiveness in Wikipedia articles, news, and scientific papers, entity set expansion in software-related texts, such as StackOverflow question-answer threads and GitHub issue reports, has been largely unexplored. Yet, there is an increasing interest in extracting software-related entities, which is a fundamental step towards software knowledge graph construction and can benefit IT operations and service management. For example, in the StackOverflowNER dataset \cite{tabassum2020code}, 20 types of entities, including \textsc{Library}, \textsc{OS}, \textsc{Application}, and \textsc{Language}, are annotated for entity-centric studies.

From the technical perspective, we identify two problems that are less concerned by existing studies: (1) \textit{How to deal with multiple types of seed entities simultaneously?} To construct a heterogeneous knowledge graph, one needs to extract multiple types of entities. If more than one type of seeds are provided for set expansion, then for a given entity type, seeds from other types can serve as negative examples and help determine the expansion boundary. For example, given two sets of seeds $\{$``\textit{Java}'', ``\textit{C++}'', ``\textit{PHP}''$\}$ and $\{$``\textit{Windows}'', ``\textit{iOS}'', ``\textit{Ubuntu}''$\}$, we know that the three \textsc{OS} seeds do not belong to the \textsc{Language} type, and all the entities extracted for \textsc{OS} during expansion should be far from \textsc{Language} as well. Without such guidance, the expansion process may suffer from semantic drifting and entity intrusion \cite{huang2020guiding}. (2) \textit{How to leverage the power of pre-trained language models?} Pre-trained language models (PLMs) such as BERT \cite{Devlin2019BERTPO} have achieved significant performance improvement in a wide spectrum of text mining tasks by learning contextualized word embeddings. The generic knowledge learned by PLMs from web-scale corpora may complement the signals we can obtain from the input corpus. For example, for some less popular programming languages such as ``\textit{Kotlin}'' and ``\textit{Groovy}'', their occurrences may not be very frequent in the input corpus, in which case their semantics cannot be accurately learned solely from local contexts. In comparison, PLMs may have learned some knowledge of them from their Wikipedia pages during pre-training.

\vspace{1mm}

\noindent \textbf{Contributions.} In this paper, we aim to tackle the aforementioned two problems of entity set expansion and apply our framework to the StackOverflow domain. To be specific, first, we study the task of \textit{entity set co-expansion}, which takes multiple types of seed entities as input and expands them simultaneously. We propose a framework, called \textsc{\model}, that iteratively expands each entity set while keeping mutual exclusivity of all types. Second, to utilize PLMs, we feed sentences containing the candidate entities into BERTOverflow \cite{tabassum2020code} to 
derive entity representations based on both the entity themselves and their contexts. The obtained representations then play a key role in calculating similarities between entities during our co-expansion process. We conduct experiments on four entity types -- \textsc{Library}, \textsc{OS}, \textsc{Application}, and \textsc{Language} -- from the StackOverflowNER dataset \cite{tabassum2020code}, and show that our proposed \textsc{\model} framework outperforms strong baselines including SetExpan \cite{shen2017setexpan} and CGExpan \cite{zhang2020empower}.
\section{Related Work}
There have been many studies on entity set expansion. For a more detailed review of related work, please refer to \cite{shen2022automated}. Early studies such as EgoSet \cite{rong2016egoset} and SetExpan \cite{shen2017setexpan} iteratively bootstrap the entity set by selecting skip-gram features and ranking new entities. Later, SetExpander \cite{mamou2018term} and CaSE \cite{yu2019corpus} propose to capture distributional similarity between words during expansion based on context-free word embeddings. More recently, CGExpan \cite{zhang2020empower} enhances entity set expansion with language model probing. In contrast, our \textsc{\model} framework utilizes PLMs in a different way by obtaining contextualized embeddings of candidate entities.
\section{Problem Definition}
Our task is formally defined as follows.

\vspace{1mm}

\noindent \textsc{Definition 1. (Entity Set Co-Expansion)} 
\textit{Given a corpus $\mathcal{D}$ and multiple entity sets $\mathcal{E}_1,\mathcal{E}_2,...,\mathcal{E}_M$ describing $M$ different entity types, where each entity set contains several (e.g., 5-10) seed entities belonging to one entity type (i.e., $\mathcal{E}_i=\{e_{i,1},...,e_{i,N}\}$), our task is to extract a set of new entities $\widetilde{\mathcal{E}}_i=\{e_{i,N+1},...,e_{i,N+K}\}$ from $\mathcal{D}$ for each entity type.}

To perform entity set co-expansion, one needs to first extract a candidate pool $\mathcal{P}$ of noun phrases from $\mathcal{D}$, which can be done by applying common phrase mining tools \cite{shang2018automated}.

\section{The \textsc{\model} Framework}

In this section, we first discuss how we utilize a large pre-trained language model to get entity embeddings. Then, we will introduce our new entity set co-expansion method.

\subsection{Entity Embeddings with Masked Language Model}

We first use a PLM to get entity embeddings based on the context information provided in the corpus. The masked language modeling (MLM) task is first proposed in BERT \cite{Devlin2019BERTPO} as a training objective for large PLMs. Basically, it is a cloze-filling task such that the model is trained to recover some tokens that are randomly selected and replaced by the special \texttt{[MASK]} token in the input sentences. Such a model after pre-training shows superior language representation power by capturing contextualized information.

However, PLMs can only embed single tokens in its fixed vocabulary, while entities are often multi-token phrases (e.g., ``\textit{Windows XP}''). Therefore, we introduce two strategies to get entity embeddings with PLMs \cite{Zhang2022EvMine,zhang2022seed}. (1) Given an entity $e$ and a sentence $s$ containing it, we can get its \emph{content embedding} $\mathbf{h}^{\text{content}}_{e|s}$ by feeding the original sentence into a PLM. Since the entity may be tokenized into multiple tokens by the model (e.g., ``\textit{Windows}'' and ``\textit{XP}''), we take the average of all its corresponding token's output embeddings as its context embedding for this sentence. (2) Then, for the same entity and sentence, we can also get its \emph{context embedding} $\mathbf{h}^{\text{context}}_{e|s}$ by replacing the entire entity with the \texttt{[MASK]} token. Then, the output embedding of the \texttt{[MASK]} token is used as its context embedding, because PLMs can only see its surrounding context to infer its semantic. 

Since the above strategies can only get entity embeddings based on one sentence, we further leverage the corpus to get corpus-level entity embeddings. For each candidate entity $e$, we find the complete set $\mathcal{S}_e$ of sentences containing it from the corpus. Then, we get its content and context embeddings for each sentence and then take the average over all sentences to get the corpus-level embeddings.
\begin{align*}
    \mathbf{h}^{\text{content}}_{e} &= \frac{1}{\abs{\mathcal{S}_e}} \sum_{s \in \mathcal{S}_e} \mathbf{h}^{\text{content}}_{e|s}, \\
    \mathbf{h}^{\text{context}}_{e} &= \frac{1}{\abs{\mathcal{S}_e}} \sum_{s \in \mathcal{S}_e} \mathbf{h}^{\text{context}}_{e|s}.
\end{align*}
We can also get a third type of embeddings by concatenating $\mathbf{h}^{\text{content}}_{e}$ and $\mathbf{h}^{\text{context}}_{e}$ to capture both signals.
\begin{equation*}
    \mathbf{h}^{\text{both}}_{e} = [\mathbf{h}^{\text{content}}_{e};\mathbf{h}^{\text{context}}_{e}].
\end{equation*}
We will study the effects of using each type of embeddings for the set expansion task in the experiments.

\subsection{Iterative Entity Set Co-Expansion}

After getting the PLM-based embeddings for each entity, $\mathbf{h}^X_e$ \ $(X \in \{\text{content}, \text{context}, \text{both}\})$, we propose an entity set co-expansion method based on the embeddings.

We first define a similarity score between an entity set $\mathcal{E}$ and a candidate entity $e$ as the average of cosine similarity between $e$ and each entity currently in $\mathcal{E}$. That is
\begin{equation*}
    \text{sim}(e, \mathcal{E}) = \frac{1}{\abs{\mathcal{E}}} \sum_{e' \in \mathcal{E}} \cos(\mathbf{h}^X_e, \mathbf{h}^X_{e'}).
\end{equation*}

Because we have multiple entity sets to expand simultaneously, for one target set, the remaining ones from different semantic classes can serve as its negative examples (i.e., irrelevant entities) to guide the expansion process. To be specific, after calculating the similarity score between each pair of candidate entity and entity set, we compare the scores for each candidate entity and select the set with the maximum score, which we name \textit{the matched set of an entity}. For example, if a candidate entity has a similarity score of 0.7 with expanded \textsc{Library} entities and a score of 0.3 with all other entity sets, we will view the \textsc{Library} entity set as its matched set and \textsc{OS}, \textsc{Application}, and \textsc{Language} entities as irrelevant ones. Formally, given a candidate entity $e$ and all entity sets to expand $\mathcal{E}_1, \mathcal{E}_2, ..., \mathcal{E}_M$, the matched set of $e$ is
\begin{equation*}
    \mathcal{E}^*_{e} = \argmax_{\mathcal{E}_i, i \in \{1, ..., M\}} \text{sim}(e, \mathcal{E}_i).
\end{equation*}
By doing so, each entity will only be expanded to the set with the highest score.
Note that for ambiguous entities that are relevant to more than one type, its highest and second highest scores may be close, in which case directly picking the highest one can be risky. To tackle this problem, we propose an iterative framework. In each iteration, only very top-ranked candidate entities will be expanded. Since it is difficult to simultaneously achieve very high similarities with more than one type, ambiguous entities are unlikely to be expanded under this strategy.

The complete set co-expansion method is as follows: given all the current entity sets and the candidate entities, we first calculate the similarity score for each pair of candidate entity and entity set and find the matched set of each entity as defined above. Then, we will select top-$k$ (e.g., $k=10$) entities $\mathcal{T}^k \subseteq \mathcal{P}$ with the highest scores to their matched sets.
\begin{equation*}
    \mathcal{T}^k = \argmax_{\mathcal{T} \subseteq \mathcal{P}, |\mathcal{T}|=k} \sum_{e \in \mathcal{T}} \text{sim}(e, \mathcal{E}^*_{e}).
\end{equation*}
Each entity of $\mathcal{T}^k$ will then be used to expand its matched set. After expanding these $k$ entities, we can re-calculate the entity-entity set scores with the updated sets and repeat the entity selection process, which becomes an iterative expansion framework. Finally, the expansion process will stop after all sets reach a given target size $t$.
\section{Experiments}

\begin{table}
\centering
\small
\caption{Selected seeds for each type.}
\scalebox{0.9}{
\begin{tabular}{ll}
\toprule
\textbf{Type}  & \textbf{Seeds}  \\ 
\midrule
\textsc{Library} & \mquote{jquery}, \mquote{api}, \mquote{angular},\\& \mquote{django}, \mquote{spring}\\
\midrule
\textsc{OS} & \mquote{windows}, \mquote{android}, \mquote{linux}, \\& \mquote{ios}, \mquote{ubuntu}\\
\midrule
\textsc{Application} & \mquote{browser}, \mquote{mysql}, \mquote{git},\\& \mquote{chrome}, \mquote{excel}, \mquote{visual studio}\\
\midrule
\textsc{Language} & \mquote{javascript}, \mquote{java}, \mquote{php}, \mquote{html}, \\& \mquote{c++}, \mquote{sql}, \mquote{python}\\
\bottomrule
\end{tabular}
}
\vspace{-0.2cm}
\label{table:seeds}
\end{table}

\subsection{Dataset}
Our corpus $\mathcal{D}$ consists of two parts: one is the StackOverflowNER dataset \cite{tabassum2020code} (but we do not use the annotations inside as supervision), and the other is a sampled subcorpus of the Stack Exchange data dump\footnote{\url{https://archive.org/download/stackexchange/stackoverflow.com-Posts.7z}} which has $\sim$1.26M questions and answers.

We consider four entity types -- \textsc{Library}, \textsc{OS}, \textsc{Application}, and \textsc{Language}. For each type, we select 5-7 seed entities which are the most frequently annotated ones in StackOverflowNER. The selected seeds are listed in Table \ref{table:seeds}.

To get the candidate entity pool from the corpus, we apply a phrase mining tool, AutoPhrase \cite{shang2018automated}, to first get all quality phrases. Then, we use spaCy\footnote{\url{https://spacy.io}} to only keep those noun phrases. Finally, we get 48,178 entities in the candidate pool.

\subsection{Compared Methods}
We compare the following entity set expansion methods.
\begin{itemize} 
    \item \textbf{SetExpan} \cite{shen2017setexpan}: This method iteratively expands the entity sets by selecting skip-gram context features and scoring entities with a rank ensemble method.
    \item \textbf{CGExpan} \cite{zhang2020empower}: This method uses a pre-trained language model to predict the type name of each entity set and use the name to guide the expansion process.
    \item \textbf{\textsc{\model}-content}: This is a variant of our proposed method using \emph{content} embeddings only.
    \item \textbf{\textsc{\model}-context}: This is a variant of our proposed method using \emph{context} embeddings only.
    \item \textbf{\textsc{\model}-both}: This is a variant of our proposed method using concatenations of \emph{both} embeddings.
\end{itemize}

\subsection{Evaluation Metric}
We use Precision@$K$ (P@$K$) as our evaluation metric. To be specific, for each entity type, we check how many of the top-$K$ extracted new entities $\widetilde{\mathcal{E}}_i=\{e_{i,N+1},...,e_{i,N+K}\}$ belong to the same entity type as $\mathcal{E}_i$. Then, we compute the average proportions across all entity types. Formally, if we use $e \sim \mathcal{E}_i$ to denote that the entity $e$ has the same type as the seeds in $\mathcal{E}_i$, then P@$K$ can be defined as
\begin{equation}
    {\rm P@}K = \frac{1}{M}\sum_{i=1}^M\frac{1}{K}\sum_{j=1}^K {\bf 1}(e_{i, N+j}\sim \mathcal{E}_i), \notag
\end{equation}
where ${\bf 1}(\cdot)$ is the indicator function.

\subsection{Hyperparameters and Implementation}

We use BERTOverflow \cite{tabassum2020code} as the PLM to get entity embeddings, which is a BERT-base model fine-tuned on the StackOverflowNER corpus. The target expansion size $t$ is set to 30. In each iteration of \textsc{\model}, top-10 (i.e., $k=10$) entities are selected to expand the entity sets before re-calculating the similarity scores in the next iteration.

\subsection{Performance Comparison}
\begin{table}
\centering
\small
\caption{P@$k$ scores of all compared methods.}
\scalebox{0.9}{
\begin{tabular}{lcccccc}
\toprule
\textbf{Methods}  & \textbf{P@10}  & \textbf{P@20} & \textbf{P@30} \\ 
\midrule
SetExpan~\cite{shen2017setexpan} & 0.325 & 0.350 & 0.358\\ 
CGExpan~\cite{zhang2020empower} & 0.775 & 0.725 & 0.708\\
\midrule
\textsc{\model}\\
\quad-content & 0.825 & 0.763 & 0.692\\
\quad-context & 0.825 & \textbf{0.850} & \textbf{0.842}\\
\quad-both & \textbf{0.925} & 0.837 & 0.742\\
\bottomrule
\end{tabular}
}
\vspace{-0.2cm}
\label{table:results}
\end{table}

Table \ref{table:results} shows the P@$K$ scores of all compared methods, where $K=$ 10, 20, and 30. From Table \ref{table:results}, we can observe that: (1) \textsc{\model}-context and \textsc{\model}-both consistently and significantly outperform the baselines, indicating the effectiveness of our proposed framework. Using skip-gram features only, SetExpan performs not so well. Enhanced by the power of PLMs, CGExpan achieves much better performance than SetExpan, but still underperforms \textsc{\model} in most cases. This is mainly because CGExpan does not have a specific design to expand multiple types of entities simultaneously. (2) Among the three variants of \textsc{\model}, \textsc{\model}-both has the highest P@10 score, while \textsc{\model}-context has the highest P@20 and P@30 scores. This observation implies that content information can only benefit the precision of top-ranked entities, while context information is more useful for extracting accurate lower-ranked entities. We will explain the reason for this through a case study.

\subsection{Case Study}
\begin{table*}
\small
\centering
\caption{Expanded entity sets for \textsc{OS} and \textsc{Language} types, with erroneous entities colored {\red red} .}
\scalebox{0.9}{
\begin{tabular}{c|c|c|c|c|c|c|c|c|c}
\toprule
Entity Type & Seed Entity Set & \multicolumn{2}{c|}{SetExpan} & \multicolumn{2}{c|}{\textsc{\model}-both} & \multicolumn{2}{c|}{\textsc{\model}-content} & \multicolumn{2}{c}{\textsc{\model}-context} \\ \midrule
                && 1  & \mquote{window}  & 1  & \mquote{ms windows}  & 1  & \mquote{microsoft window} & 1  & \mquote{macos}   \\ 
                && 2  & \red\mquote{iphone} & 2  & \mquote{microsoft window}  & 2  & \mquote{gnu linux} & 2  & \mquote{osx}  \\ 
                && 3  & \red\mquote{a} & 3  & \mquote{gnu linux} & 3  & \mquote{window server} & 3  & \mquote{macosx}  \\ 
                && 4  & \red\mquote{mac} & 4  & \mquote{window ce} & 4  & \mquote{windows server} & 4  & \red\mquote{mac}  \\ 
                && 5  & \red\mquote{local} & 5  & \mquote{arch linux} & 5  & \mquote{ms windows} & 5  & \mquote{mac osx}  \\ 
                &&    & ... &    & ...  &    & ...  &    & ...                                     \\ 
                && 26 & \red\mquote{modern}  & 26 & \mquote{window server 2008} & 26 & \mquote{window 2008 server} & 26 & \mquote{debian jessie} \\ 
                && 27 & \red\mquote{office}  & 27 & \red\mquote{window powershell} & 27 & \red\mquote{windows api} & 27 & \mquote{windows vista} \\ 
                && 28 & \red\mquote{regular}  & 28 & \mquote{window phone}  & 28 & \red\mquote{window explorer}  & 28 & \red\mquote{nix}   \\ 
                && 29 & \red\mquote{4 gb} & 29 & \mquote{windows 8}  & 29 & \mquote{window mobile}  & 29 & \mquote{windows 7} \\ 
                \multirow{-11}{*}{\textsc{OS}}&
                \multirow{-11}{*}{\begin{tabular}[c]{@{}c@{}}\{\mquote{windows}, \\ \mquote{android}, \\ \mquote{linux}, \\ \mquote{ios}, \\ \mquote{ubuntu}\}\end{tabular}} 
                & 30 & \red\mquote{remote} & 30 & \red\mquote{windows powershell} & 30 & \mquote{windows 8} & 30 & \mquote{window vista} \\
\midrule
                && 1  & \mquote{c}  & 1  & \mquote{c}  & 1  & \mquote{objective c}  & 1  & \mquote{c}   \\ 
                && 2  & \mquote{js} & 2  & \mquote{objective c}  & 2  & \mquote{obj c}  & 2  & \mquote{vb}  \\ 
                && 3  & \mquote{css} & 3  & \mquote{vb net} & 3  & \mquote{c} & 3  & \mquote{go}  \\ 
                && 4  & \mquote{ruby} & 4  & \mquote{obj c} & 4  & \mquote{c sharp} & 4  & \mquote{golang}  \\ 
                && 5  & \red\mquote{korean} & 5  & \mquote{c sharp} & 5  & \mquote{turbo c} & 5  & \mquote{elixir}  \\ 
                &&    & ... &    & ...  &    & ...  &    & ...                                     \\ 
                && 26 & \red\mquote{a}  & 26 & \red\mquote{asp net webapi} & 26 & \red\mquote{java net urlconnection} & 26 & \mquote{sml} \\ 
                && 27 & \mquote{haskell}  & 27 & \red\mquote{dot net} & 27 & \red\mquote{js main js} & 27 & \mquote{racket} \\ 
                && 28 & \mquote{clojure}  & 28 & \red\mquote{asp net core webapi}  & 28 & \red\mquote{discord js}  & 28 & \mquote{vhdl}   \\ 
                && 29 & \mquote{powershell} & 29 & \red\mquote{dot net core}  & 29 & \red\mquote{tcp ip}  & 29 & \mquote{nim} \\ 
                \multirow{-11}{*}{\textsc{Language}}&
                \multirow{-11}{*}{\begin{tabular}[c]{@{}c@{}}\{\mquote{javascript}, \\ \mquote{java}, \\ \mquote{php}, \\ \mquote{html}, \\ \mquote{c++}, \\ \mquote{sql}, \\ \mquote{python}\}\end{tabular}} 
                & 30 & \mquote{vb net} & 30 & \red\mquote{asp net boilerplate} & 30 & \red\mquote{crypto js} & 30 & \mquote{scala} \\
\bottomrule
\end{tabular}
}
\label{table:cases} 
\vspace{-0.2cm}
\end{table*}

Table \ref{table:cases} shows the expanded entity sets of SetExpan, \textsc{\model}-both, \textsc{\model}-content, and \textsc{\model}-context for two types, where we list both top-ranked entities (i.e., 1$^{\rm st}$ to 5$^{\rm th}$) and lower-ranked ones (i.e., 26$^{\rm th}$ to 30$^{\rm th}$). We can see that: (1) SetExpan picks many general terms (e.g., ``\textit{a}'', ``\textit{local}'', ``\textit{modern}'') that do not belong to any specific entity type. (2) For \textsc{\model}-both and \textsc{\model}-content, because we use entity content (i.e., tokens in each entity) to derive embeddings, the extracted terms are more likely to have lexical overlap with the seed entities. For top-ranked entities (e.g., ``\textit{ms windows}'', ``\textit{gnu linux}'', ``\textit{c}''), this strategy yields high precision. However, when it comes to lower-ranked entities (e.g., ``\textit{windows powershell}'', ``\textit{windows api}''), lexical overlap does not necessarily indicate that two entities belong to the same type. This explains our observation from Table \ref{table:results} that content information is not as robust as context information for extracting lower-ranked entities.
\section{Conclusions and Future Work}
In this paper, we study the problem of entity set co-expansion in StackOverflow to extract \textsc{Library}, \textsc{OS}, \textsc{Application}, and \textsc{Language} entities with just a few seeds. We propose to leverage a PLM to derive entity embeddings based on entity content and/or context. Then, an iterative co-expansion framework is proposed to simultaneously enrich multiple sets of entities based on the calculated entity embeddings. Experimental results show that our proposed \textsc{\model} framework significantly outperforms strong baselines such as SetExpan and CGExpan. Through quantitative and qualitative analyses, we also conclude that context signals are more robust than content signals for extracting lower-ranked entities.
For future work, first, it is of our interest to generalize our framework to more types of entities such as \textsc{Version} and \textsc{Device}.
Second, we would like to explore the possibility of applying our entity set co-expansion results to distantly supervised or few-shot named entity recognition in StackOverflow and GitHub issue reports.

\section*{Acknowledgments}
We thank anonymous reviewers for their valuable and insightful feedback. This work was supported by the IBM-Illinois Discovery Accelerator Institute and National Science Foundation IIS-19-56151, IIS-17-41317, and IIS 17-04532.

\bibliographystyle{abbrv}
\bibliography{bigdata22}

\end{spacing}

\end{document}